\pdfoutput=1
\documentclass[11pt,a4paper]{article}
\usepackage{arxiv}
\usepackage[utf8]{inputenc} 
\usepackage[T1]{fontenc}    
\usepackage{amsfonts}       
\usepackage{nicefrac}       
\usepackage{lipsum}		
\usepackage{natbib}
\usepackage{doi}

\usepackage{hyperref}
\usepackage{times}
\usepackage{latexsym}
\usepackage{xspace}
\usepackage{multirow}
\usepackage{url}
\usepackage{booktabs}
\usepackage{tikz,tikz-qtree}
\usepackage{pgfplots}
\usepackage{amssymb}
\usepackage{xfrac}
\usepackage{graphicx}
\usepackage{tablefootnote}
\usepackage{amsmath}
\usepackage{enumitem}
\usepackage{todonotes}
\usepackage{subfigure}
\pgfplotsset{compat=1.14}

\usepackage{algorithm}
\usepackage{algorithmic}

\usepackage{microtype}

\relax
\def\red#1{{\color{red}{#1}}}

\definecolor{g-red}{HTML}{DB4437}
\definecolor{g-blue}{HTML}{4285F4}
\definecolor{g-green}{HTML}{0F9D58}
\definecolor{g-yellow}{HTML}{F4B400}
\definecolor{g-orange}{HTML}{FF9800}
\definecolor{g-grey}{HTML}{9E9E9E}
\renewcommand{\v}[1]{\boldsymbol{#1}}
\renewcommand{\t}[1]{\text{#1}}

\DeclareMathOperator*{\softmax}{softmax}
\DeclareMathOperator*{\sigmoid}{sigmoid}
\DeclareMathOperator{\relu}{ReLU}

\setenumerate[1]{itemsep=0pt,partopsep=0pt,parsep=\parskip,topsep=5pt}
\setitemize[1]{itemsep=0pt,partopsep=0pt,parsep=\parskip,topsep=5pt}
\setdescription{itemsep=0pt,partopsep=0pt,parsep=\parskip,topsep=5pt}

\title{AdvPicker: Effectively Leveraging Unlabeled Data via Adversarial Discriminator for Cross-Lingual NER}
\renewcommand{\shorttitle}{AdvPicker: Effectively Leveraging Unlabeled Data via Adversarial Discriminator...}

\author{Weile Chen$^1$, Huiqiang Jiang$^2$, Qianhui Wu$^3$, B{\"o}rje F. Karlsson$^4$, and Yi Guan$^1$ \\
    $^1$Harbin Institute of Technology, Harbin, 15000, China\\
    $^2$Peking University, Beijing, 100871, China\\
	$^3$Department of Automation, Tsinghua University, Beijing, 100084, China \\
	$^4$Microsoft Research, Beijing, 100080, China \\
	\texttt{chen.weile7@gmail.com}, \texttt{jhq@pku.edu.cn}, \texttt{guanyi@hit.edu.cn}, \\
	\texttt{wuqianhui@tsinghua.org.cn}, \texttt{borje.karlsson@microsoft.com}
}

\date{}

\begin{document}
\maketitle
\begin{abstract}
    Neural methods have been shown to achieve high performance in Named Entity Recognition (NER), but rely on costly high-quality labeled data for training, which is not always available across languages.
While previous works have shown that unlabeled data in a target language can be used to improve cross-lingual model performance, we propose a novel adversarial approach (\textit{AdvPicker}) to better leverage such data and further improve results.
We design an adversarial learning framework in which an encoder learns entity domain knowledge from labeled source-language data and better shared features are captured via adversarial training - where a discriminator selects less language-dependent target-language data via similarity to the source language.
Experimental results on standard benchmark datasets well demonstrate that the proposed method benefits strongly from this data selection process and outperforms existing state-of-the-art methods; without requiring any additional external resources (e.g., gazetteers or via machine translation). \footnote{Code is publicly available at \url{https://aka.ms/AdvPicker}}
\end{abstract}
\section{Introduction}
\label{sec:Introduction}
Named entity recognition (NER) is a fundamental information extraction task, which seeks to identify named entities in text and classify them into pre-defined entity types (such as person, organization, location, etc.) and it is key in various downstream tasks, e.g., question answering \cite{molla-etal-2006-named}.
Neural NER models are highly successful for languages with a large amount of quality annotated data.
However, most languages don't have enough labeled data to train a fully supervised model. 
This motivates research on cross-lingual transfer, which leverages labeled data from a source language (e.g., English) to address the lack of training data problem in a target language. 
In this paper, following \citet{wu-dredze-2019-beto} and \citet{wu2020unitrans}, we focus on zero-shot cross-lingual NER, where labeled data is not available in the target language. 

The state-of-the-art methods for zero-shot cross-lingual NER are mainly divided into three categories: i) feature-based methods \cite{wu-dredze-2019-beto,wu2020enhanced,pfeiffer-etal-2020-mad}, which train a NER model to capture language-independent features of the labeled source-language data and then apply it to the target language; ii) translation-based methods \cite{mayhew2017cheap,xie-etal-2018-neural}, which build pseudo target-language dataset via translating from labeled source-language data and mapping entity labels; and iii) pseudo-labeling methods, which generate pseudo-labeled data for training a target-language NER model via a source-language model \cite{wu2020unitrans} or annotation projection \cite{ni2017weakly}.

However, each method has its own disadvantages. Feature-based methods only learn the knowledge in the source language, but cannot leverage any target-language information.  
Translation-based methods require high-quality translation resources, which are expensive to obtain.
And pseudo-labeled methods assume that all pseudo-labeled data is beneficial for cross-lingual transfer learning, which is not always the case.

Therefore, here we propose a novel approach -- \textit{AdvPicker} -- which combines feature-based and pseudo-labeling methods, while not requiring any extra costly resources (e.g., translation models or parallel data). Furthermore, to address the described problems, we enhance the source-language NER model with unlabeled target language data via adversarial training. 
Unlike other pseudo-labeling methods, we only leverage the language-independent pseudo-labeled data selected by an adversarial discriminator, to alleviate overfitting the model in language-specific features of the source-language.

Specifically, we first train an encoder and a NER classifier on labeled source-language data to learn entity domain knowledge. Meanwhile, a language discriminator and the encoder are trained on a token-level adversarial task which enhances the ability of the encoder to capture shared features.
We then apply the encoder and the NER classifier on unlabeled target-language data to generate pseudo-labels and use an adversarial discriminator to select less language-specific data samples. 
Finally, we utilize knowledge distillation to train a target-language NER model on this selected dataset.

We evaluate our proposed \textit{AdvPicker} over 3 target languages on standard benchmark datasets. Our experimental results show that the proposed method benefits strongly from this data selection process and outperforms existing SOTA methods;  without  requiring  any  additional external resources (e.g., gazetteers or machine translation).

Our major contributions are as follows:
\begin{itemize}
\item We propose a novel approach to combine feature-based and pseudo-labeling methods via language adversarial learning for cross-lingual NER;
\item We adopt an adversarial discriminator to select what language-independent data to leverage in training a cross-lingual NER model to improved performance. To the best of our knowledge, this is the first successful attempt in selecting data by adversarial discriminator for XL-NER;
\item Experiments on standard multi-lingual datasets showcase \textit{AdvPicker} achieves new state-of-the-art results in cross-lingual NER.
\end{itemize}

\section{Related Work}
\label{sec:Related Work}

\subsection{Cross-Lingual NER}
Cross-lingual transfer for NER has been widely studied in recent years.
Prior works are divided into three categories: feature-based, translation-based, and pseudo-labeling.

Feature-based methods generally use language-independent features to train a NER model in the labeled source-language data, which include word clusters \cite{tackstrom2012}, Wikifier features \cite{tsai2016cross}, gazetteers \cite{zirikly2015cross}, and aligned word representations \cite{ni2017weakly,wu-dredze-2019-beto}, etc. Moreover, for language-independent features, adversarial learning was applied on word/char embedding layers \cite{huang-etal-2019-cross,bari2019zero} or encoders \cite{zhou2019dual,adv-follow}.
Translation-based methods generally use pseudo target-language data translated from labeled source-language data. \citet{ni2017weakly} proposed to project labels from the source language into the target language by using word alignment information. 
Most recent methods translate the annotated corpus in the source language to the target language word-by-word \cite{xie-etal-2018-neural} or phrase-by-phrase \cite{mayhew2017cheap} and then copy the labels for each word/phrase to their translations. While \cite{jain2019entity} proposed to translate full sentences in the source language and project entity labels to target-language sentences.

To leverage unlabeled target-language data, pseudo-labeling methods generate the pseudo-labels by annotation projection on comparable corpora \cite{ni2017weakly} or via models trained on source-language labeled data \cite{wu2020unitrans}.

In this paper, we propose \textit{AdvPicker}, an approach that requires no translation and combines feature-based and pseudo-labeling methods. Moreover, we leverage pseudo-labeled data differently from other pseudo-labeling methods. Through adversarial training, we select language-independent pseudo-labeled data for training a new target-language model.

\subsection{Language Adversarial Learning}
Language-adversarial training \cite{zhang2017adversarial} was proposed for the unsupervised bilingual lexicon induction task. And it has been applied in inducing language-independent features for cross-lingual tasks in NER \cite{zhou2019dual,xie-etal-2018-neural}, text classification \cite{chen-etal-2019-multi-source}, and sentiment classification \cite{chen2018adversarial}. 

\citet{adv-follow} proposed a multilingual BERT with sentence-level adversarial learning. 
However, this method does not improve cross-lingual NER performance significantly. 
To address this limitation, \textit{AdvPicker} uses multilingual BERT with token-level adversarial training for cross-lingual NER, which induces more language-independent features for each token embedding.

\subsection{Knowledge Distillation}
Knowledge distillation was proposed to compress models \cite{bucilu2006} or ensembles of models \cite{rusu2015,hinton2015kd,sanh2019,mukherjee-hassan-awadallah-2020-xtremedistil} via transferring knowledge from one or more models (teacher models) to a smaller one (student model).
Besides model compression, knowledge distillation has also been applied to various tasks, like cross-modal learning \cite{hu2020creating}, machine translation \cite{weng2020acquiring}, and automated machine learning \cite{kang2020towards}.

In this paper, we adapt knowledge distillation to leverage unlabeled data in the cross-lingual NER task. This helps the student model learn richer information from easily obtainable data (with pseudo-labels).

\section{AdvPicker}
\label{sec:model}

\begin{figure*}[htb!]
    \begin{center}
    \includegraphics[width=1\textwidth]{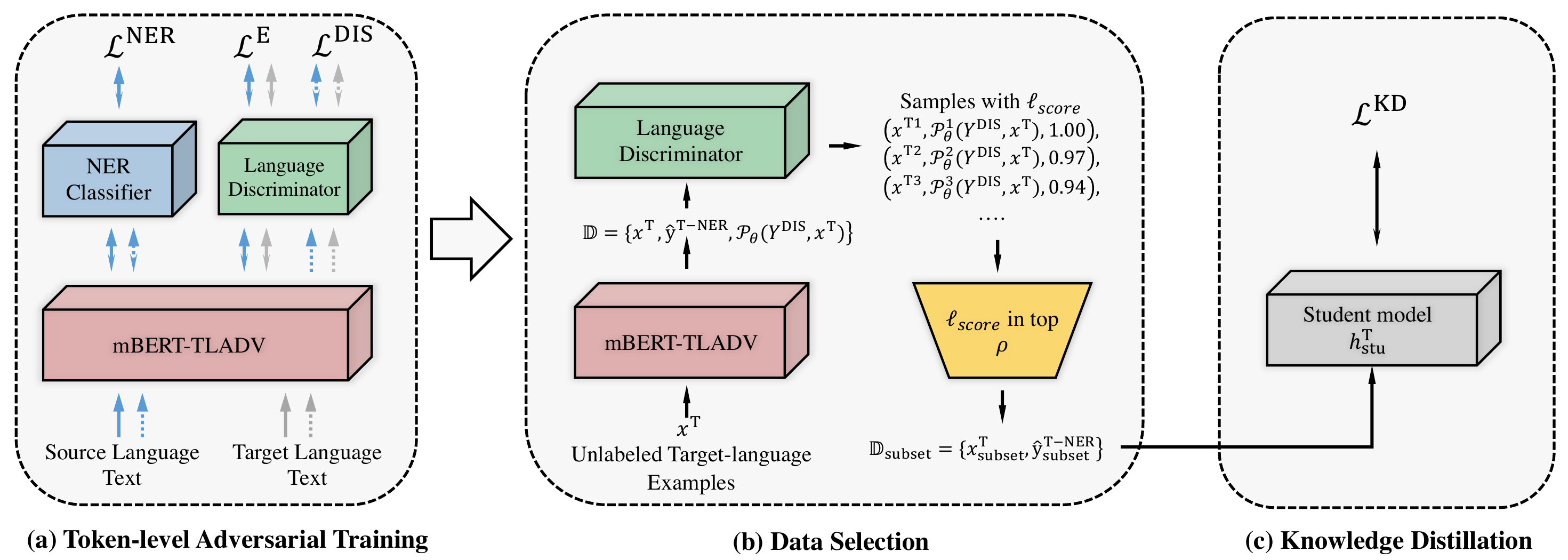}
    \caption{Framework of the proposed \textit{AdvPicker}. a) Overview of the token-level adversarial training process. The lines illustrate the training flows and the arrows indicate forward or backward propagation. Blue lines show the flow for source-language samples and grey ones are for the target language. $\mathcal L^{\t{NER}}$, $\mathcal L^{\t{E}}$, and $\mathcal L^{\t{DIS}}$ are the losses of the NER classifier, encoder and discriminator modules in \textit{AdvPicker} respectively (Section \ref{sec:tladv}). Encoder and NER classifier are trained together on source-language samples (blue solid lines on the left side). Encoder and discriminator are trained for the adversarial task (on the right side). b) Language-independent data selection on pseudo-labeled data. c) Knowledge distillation on selected data.}
    \label{fig:framework}
    \end{center}
\end{figure*}

In this section, we introduce our approach (\textit{AdvPicker}) which utilizes the adversarial learning approach to select language-independent pseudo-labeled data for training an effective target-language NER model.
Figure \ref{fig:framework} illustrates the framework of the proposed \textit{AdvPicker}. 
Specifically, as shown in Figure \ref{fig:framework}(a), we train an encoder and a NER classifier on the labeled source-language data. Meanwhile, a language discriminator and the encoder are trained on the token-level adversarial task.
We then apply encoder and classifier over unlabeled target-language data to generate pseudo-labels and use the adversarial discriminator to select the most language-independent pseudo-labeled data samples.
Finally, we utilize knowledge distillation to train a target-language NER model on this selected dataset.

In the following section, we describe the language-independent data selection process, including the token-level adversarial training, data selection by the discriminator, and knowledge distillation on select language-independent data.

\subsection{Token-level Adversarial Training for Cross-Lingual NER}
\label{sec:tladv}
To avoid the model overfitting on language-specific features of the source-language, we propose the token-level adversarial learning (TLADV) framework, which is shown in Figure \ref{fig:framework}(a).

Following \citet{adv-follow}, we formulate adversarial cross-lingual NER as a multi-task problem: i) NER  and ii) binary language classification (i.e source vs. target language). 
For the NER task, we train the encoder and classification layer on NER annotated text in the source language. The encoder learns to capture the NER features of the input sentences and then the classification layer tries to predict the entity labels for each word based on their feature vectors. 

For the language classification task, we train a language discriminator and an encoder on the labeled source-language dataset and unlabeled target-language data. 
The language discriminator is added to classify whether an embedding generated by the encoder is associated to the source or the target language. 
The encoder tries to produce language-independent embeddings that are difficult for the language discriminator to classify correctly.
We define the encoder, the language discriminator, and their objectives as follows:

\noindent \textbf{Encoder} Given an input sentence $\v{x} = [x_i]_{1 \le i \le N}$ with $N$ words, we feed it into encoder $\v E$ to obtain feature vectors $\v h = [h_i]_{1 \le i \le N}$ for all words:
\begin{equation}
    \v h = {\v{E}}(\v x)
    \label{equ:encoder}
\end{equation}
where ${\v{E}}$ is the feature encoder which generates language-independent feature vectors $\v h$ for each sentence $\v x$. Following \citet{adv-follow}, we use multilingual BERT as the feature encoder here and denote the encoder as mBERT-TLADV.

\noindent \textbf{NER Classifier} We feed $\v h$ into the NER classifier which is a linear classification layer with the $\softmax$ activation function to predict the entity label of token $\v x$.
\begin{equation}
    \mathcal{P}_{\theta}({\v Y}^{\t{NER}}) = \softmax(\v W^{\t{NER}}\v h + \v b^{\t{NER}})
    \label{equ:ner}
\end{equation}
where $\mathcal{P}_{\theta}({\v Y}^{\t{NER}}) \in \mathbb{R}^{|\mathbb{C}|}$ is the probability distribution of entity labels for token $\v x$ and $\mathbb{C}$ is the entity label set. $\v W^{\t{NER}} \in \mathbb{R}^{d_e \times |\mathbb{C}|} $ and $\v b^{\t{NER}} \in \mathbb{R}^{|\mathbb{C}|}$ denote the to-be-learned parameters with $d_e$ being the dimension of vector $\v h$.

\noindent \textbf{Language Discriminator} The language discriminator is comprised of two linear transformations and a $\relu$ function for classifying token embedding.
The $\sigmoid$ function is used to predict the probability of whether $\v h$ belongs to the source language.
\begin{equation} 
    \label{equ:discrminator}
    \mathcal{P}_{\theta}({\v Y}^{\t{DIS}}) = \sigma(\v W^{\t{DIS1}}\relu(\v W^{\t{DIS2}} \v h))
\end{equation}
where $\v W^{\t{DIS1}} \in \mathbb{R}^{d_{d} \times d_e}$ and $\v W^{\t{DIS2}} \in \mathbb{R}^{d_\ell \times d_d}$, with $d_{d}$ being the hidden dimension of discriminator and $d_\ell$ the language classification task label size. $\sigma$ is the $\sigmoid$ function to obtain the language probability of each word.

For language-adversarial training, we have 3 loss functions: the encoder loss $\mathcal L^{\t{E}}$, the language discriminator loss $\mathcal L^{\t{DIS}}$, and the NER task loss $\mathcal L^{\t{NER}}$.

\begin{algorithm}[ht]
	\caption{Pseudocode for token-level adversarial training on zero-shot cross-lingual NER.}
	\label{algo:pseudocode}
	\begin{algorithmic}[1]
	    \REQUIRE training set $\mathbb{D}_{\t{E}} = \mathbb{D}_{\t{DIS}} = \{(\v{x}, \v{y}^{\t{DIS}})\}$ and $\mathbb{D}_{\t{NER}} = \{(\v{x}, \v{y}^{\t{NER}})\}$, batch size $bs = 1$ for clarity, batch number $B$, the output probability of model $\mathcal{P}_{\theta}({\v Y}^{m})$, the hidden vector of model $\v h$.
	    \STATE Initialize hidden vector of $\v W^{*}, \v b^{*}$;
		\FOR{$i = 1, \cdots, B$}
		\FOR {model $m$ in \{$\t{NER}$, $\t{E}$, $\t{DIS}$\}}
		\STATE Sample batch $\v{(x,y)} \in \mathbb{D}_{m}$;
		\STATE \# \textit{Calculate model hidden vector};
		\STATE $\v h = \v{E}(\v{x})$, refer to Eq.(\ref{equ:encoder})
		\IF{model $m$ equal to model $\t{NER}$}
		    \STATE Calculate $\mathcal{P}_{\theta}({\v Y}^{\t{NER}})$, Eq.(\ref{equ:ner})
		\ELSE
		    \STATE  Calculate $\mathcal{P}_{\theta}({\v Y}^{\t{DIS}})$, Eq.(\ref{equ:discrminator})
		\ENDIF
		\STATE Calculate model loss $\mathcal{L}_{m}$ using $\v x$, $\v y$ and $\mathcal{P}_{\theta}({\v Y}^{{m}})$, Eq.(\ref{eq:loss-total})
		\STATE Update parameters of embeddings w.r.t. the gradients
        using $\nabla \mathcal{L}_{m}$ with parameters $\v W^{m}, \v b^{m}$ and the encoder $\v{E}$ (if $m$ in \{$\t{NER}$, $\t{E}$\}). 
        \ENDFOR
        \ENDFOR
	\end{algorithmic}
\end{algorithm}

Note that we don't add these three loss functions together for backward propagation. 
Parameters of different components in adversarial learning are alternatively updated based on the corresponding loss function, similarly to \citet{adv-follow}. 
Specifically, for the NER task, the parameters of the encoder and the NER classifier are updated based on $\mathcal L^{\t{NER}}$. 
For the adversarial task, the parameters of the encoder are updated based on $\mathcal L^{\t{E}}$, while the parameters of the discriminator are updated based on $\mathcal L^{\t{DIS}}$. Algorithm \ref{algo:pseudocode} shows the pseudocode for the adversarial training process.

\begin{equation}
\begin{aligned}
    \mathcal L^{\t{E}} = - \frac{1}{N} \sum_{i \in [1, N]} \log{\mathcal{P}_{\theta}(Y^{\t{DIS}}_{i} = {\widetilde{y}}^{\t{DIS}}_{i})} &\\ 
     + \log{\mathcal{P}_{\theta}(Y^{\t{DIS}}_{i} = {y}^{\t{DIS}}_{i})}  &\\
    \mathcal L^{\t{DIS}} = - \frac{1}{N} \sum_{i \in [1, N]} \log{\mathcal{P}_{\theta}(Y^{\t{DIS}}_{i} = y^{\t{DIS}}_{i})} &\\
     + \log{\mathcal{P}_{\theta}(Y^{\t{DIS}}_{i} = {\widetilde{y}}^{\t{DIS}}_{i})} &\\
    \mathcal L^{\t{NER}} = - \frac{1}{N} \sum_{i \in [1, N]} \log{\mathcal{P}_{\theta}(Y^{\t{NER}}_{i} = y^{\t{NER}}_{i})} &\\
    \label{eq:loss-total}
\end{aligned}
\end{equation}
where $\v{x}$ is the sentence, $\v{y}^{\t{DIS}} \in \{0, 1\}$ is the ground truth label for the language classification task, $\v{\widetilde{y}}^{\t{DIS}} \in \{0, 1\}$ is the negative label for the language classification task, and $\v{y}^{\t{NER}} \in \mathbb{R}^{N \times |\mathbb{C}|}$ is the ground truth of named entity recognition task for corresponding input $\v{x}$.

\subsection{Language-Independent Data Selection}
\label{sec:DataSelection}

To obtain the pseudo labels $\v{\hat{y}}^{\t{T-NER}}$ for target-language examples, we apply the learned mBERT-TLADV model on the unlabeled target-language data $\v x^{\t T}$. However, the pseudo-labeled dataset $\mathbb{D}=\{\v x^{\t T},\v{\hat{y}}^{\t{T-NER}}\}$ may then contain lots of language-specific samples.
We then leverage the adversarial discriminator to select pseudo language-independent samples from the generated set.

The language discriminator tries to make the encoder unable to distinguish the language of a token through confrontation. In this way, the encoder should pay more attention to features that are less related to the source language when learning the NER task.
After adversarial training, the language discriminator can still correctly classify certain embeddings with a high probability. We define these as \textit{language-specific} samples.
Other samples are ambiguous regarding language (for example, sentences with probability close to 0.5), and they are defined as samples that are more \textit{language-independent}.

In order to quantify the language independence of each sample, we use the language discriminator to calculate the probability of whether the sentence $\v{x}^{\t T}$ is from the source language $\mathcal{P}_{\theta}({\v Y}^{\t{DIS}}, \v x^{\t T})$, the formula is as follows:
\begin{equation}
    \mathcal{P}_{\theta}({\v Y}^{\t{DIS}}, \v x^{\t T}) = \sigma(\v W^{\t{DIS1}}\relu(\v W^{\t{DIS2}} \v h^{\t T}))
\end{equation}

where $\v x^{\t T}$ and $\v h^{\t T}$, respectively, denote the target-language sentence and its feature vector, and $\mathcal{P}_{\theta}({\v Y}^{\t{DIS}}, \v x^{\t T})$ is the probability of $\v x^{\t T}$ mentioned in Eq.( \ref{equ:discrminator}).

In order to select from the pseudo-labeled data, we design an index $\ell_{\t{score}}$ to represent the language independence of a sentence $\v{x}^{\t T}$ (the degree of model confusion on different languages).
We assume it follows a uniform distribution and reaches its maximum when $\mathcal{P}_{\theta}({\v Y}^{\t{DIS}}, \v x^{\t T})=0.5$. Conversely, the index is at its minimal value when $\mathcal{P}_{\theta}({\v Y}^{\t{DIS}}, \v x^{\t T})$ is equal to 0 or 1.
\begin{equation}
 \ell_{\t{score}}(\v{x}^{\t T}) = 1- \left\|\mathcal{P}_{\theta}({\v Y}^{\t{DIS}}, \v x^{\t T}) - 0.5\right\|   
\end{equation}

We select target-language samples with the highest $\ell_{\t{score}}$ in the top $\rho$ as language-independent data.
$\rho$ is a hyper-parameter which is the data ratio of pseudo-labeled data.
Finally, we obtain the selected target-language pseudo-labeled dataset $\mathbb{D}_{\t{subset}} = \{x^{\t T}_{\t{subset}}, \v{\hat{y}}^{\t{T-NER}}_{\t{subset}}\}$, a subset of target language pseudo-labeled dataset.

There are two reasons for selecting language-independent samples by the language discriminator.
First, these samples' feature vectors contain less language-specific information which is helpful for cross-lingual transfer learning.
Second, the NER classifier is trained on source-language labeled data. Therefore, it is more likely to generate high-quality predictive labels on selected target-language samples that have similar feature vectors to source-language samples.

\subsection{Knowledge Distillation on Language-Independent Data}
To leverage such less language-dependent data, we train a target-language NER model on the selected pseudo-labeled data $\mathbb{D}_{\t{subset}}$.
Considering a lot of helpful information can be carried in soft targets instead of hard targets \cite{hinton2015kd}, we use the soft labels of the selected pseudo-data to train a student model $\v h^{\t T}_{\t{stu}}$ via knowledge distillation.
To construct the student model, we used the pre-trained cased multilingual BERT (mBERT) \cite{mbert} as the initialization and a linear layer with $\softmax$ function:
\begin{equation}
    \mathcal{P}_{\theta}({\v Y}^{\t{T-NER}}) = \softmax(\v W^{\t{T-NER}}\v h^{\t T}_{\t{stu}} + \v b^{\t{T-NER}})
    \label{equ:p_stu}
\end{equation}
where $\mathcal{P}_{\theta}({\v Y}^{\t{T-NER}})$ is the distribution of entity labels probability output from the student model. $\v W^{\t{T-NER}} \in \mathbb{R}^{d_s \times |\mathbb{C}|} $ and $\v b^{\t{T-NER}} \in \mathbb{R}^{|\mathbb{C}|}$ are learnable parameters of the student NER model.

Following \citet{wu2020unitrans}, the loss function $\mathcal L^{\t{KD}}$ is defined as the \textit{mean squared error} (MSE) between the prediction output $\mathcal{P}_{\theta}({\v Y}^{\t{T-NER}})$ and the soft labels of the selected data, which is formulated as:
\begin{equation}
    \mathcal L^{\t{KD}} = \frac{1}{N} \sum_{i \in [1, N]} (\mathcal{P}_{\theta}({\v Y}^{\t{T-NER}}) - \v{\hat{y}}^{\t{T-NER}}_{\t{subset}}) ^ 2 \\
    \label{eq:loss-each}
\end{equation}
where $\v{\hat{y}}^{\t{T-NER}}_{\t{subset}} \in \mathbb{D}_{\t{subset}}$ are the selected soft labels with $N$ tokens and $\mathcal{P}_{\theta}({\v Y}^{\t{T-NER}})$ is the prediction probability of the selected sentence $\v{x}^{\t T}$. By minimizing the MSE loss, the student model is trained supervised on the target-language selected data pseudo-labels.

For inference in the target language, we only apply the student model on test cases to predict the probability distribution of entity labels for each token in sentences, as Eq.(\ref{equ:p_stu}).
To ensure the entity labels follow the NER tagging scheme, the prediction result is generated by Viterbi decoding \cite{chen2019grn}.

\section{Experiments}
\label{sec:Experiments}

\subsection{Experiment Settings}
\label{sec:ExpSetting}

\subsubsection{Datasets}
\label{sec:Datasets}
\begin{table*}[!t]
   \centering
   \vskip 0.1in
   \begin{tabular}{c|c|c|c|c}
   \toprule
      \textbf{Language} & \textbf{Type} & \textbf{Train} & \textbf{Dev} & \textbf{Test}\\ \midrule
	English [en] & \# of Sentence & 14,987 & 3,466 & 3,684  \\
	(CoNLL-2003) & \# of Entity & 23,499 & 5,942 & 5,648  \\
	\hline    
	German [de] & \# of Sentence & 12,705 & 3,068 & 3,160  \\
	(CoNLL-2003) & \# of Entity & 11,851 & 4,833 & 3,673  \\
	\hline    
	Spanish [es] & \# of Sentence & 8,323 & 1,915 & 1,517  \\
	(CoNLL-2002) & \# of Entity & 18,798 & 4,351 & 3,558  \\
	\hline    
	Dutch [nl] & \# of Sentence & 15,806 & 2,895 & 5,195  \\
	(CoNLL-2002) & \# of Entity & 13,344 & 2,616 & 3,941  \\
    \bottomrule
   \end{tabular}%
   \caption{Statistics of the benchmark datasets.}
   \label{tab:DatasetsStat}
\end{table*}
   
We conduct experiments in 4 different languages: English [en], Spanish [es], Dutch [nl], German [de].
Spanish and Dutch data is from the CoNLL-2002 NER shared task \cite{conll2002}\footnote{https://www.clips.uantwerpen.be/conll2002/ner/}, while English and German are from CoNLL-2003 \cite{conll2003}\footnote{https://www.clips.uantwerpen.be/conll2003/ner/}.
Table \ref{tab:DatasetsStat} presents some basic statistics of the datasets used in our experiments.

CoNLL-2002/2003 data uses gold standard labelling and it is tagged with four entity types: PER, LOC, ORG, and MISC. Following \citet{wu-dredze-2019-beto}, we use the BIO labeling scheme \citep{farber-etal-2008-improving} and the official split of train/validation/test sets.
As previous works \cite{tackstrom2012,jain2019entity,wu2020enhanced}, for all experiments, we always use English as source language and the others as target languages. 
Our models are trained on the training set of English and evaluated on the test sets of each target language.

Note that for each target language, we only use text in its training set to train our model with these unlabeled target language data.
In adversarial learning, we randomly sample data from all target languages and construct a target-language dataset of the same size as the English training dataset.

\subsubsection{Implementation Details}
\label{sec:Implementation}
We implement AdvPicker using PyTorch 1.6.0.
For data pre-processing, we leverage WordPiece \cite{wu2016google} to tokenize each sentence into a sequence of sub-words which are then fed into the model.
For the encoder (i.e. $E$ in Eq.(\ref{equ:encoder})) and student model (i.e. $h^{\t T}_{\t{stu}}$ in Eq.(\ref{equ:p_stu})), we employ the pre-trained cased multilingual BERT in HuggingFace’s Transformers \citep{wolf-etal-2020-transformers}\footnote{https://github.com/huggingface} as backbone model, which has 12 transformer blocks, 12 attention heads, and 768 hidden units.

We empirically select the following hyper-parameters.
Specifically, referring to the settings of \citet{wu2020enhanced}, we freeze the parameters of the embedding layer and the bottom three layers of the multilingual BERT used in the encoder and the NER student model.
We train all models using a batch size of 32, maximum sequence length of 128, a dropout rate of 0.1, and use AdamW \cite{adamw} as optimizer. For sequence prediction, we apply Viterbi decoding \cite{chen2019grn} on all models in our experiments.

Following \citet{keung-etal-2020-dont}, in all experiments the other hyper-parameters are tuned on each target language dev set.
We train all models for 10 epochs and choose the best model checkpoint with the target dev set.
For adversarial learning, we set the learning rate of 6e-5 for the NER loss $\mathcal L^{\t{NER}}$ and 6e-7 for both loss encoder $\mathcal L^{\t{E}}$ and discriminator loss $\mathcal L^{\t{DIS}}$.
For knowledge distillation, we use a learning rate of 6e-5 for the student models.
We set the hidden dimension of the discriminator as 500. For data selection, $\rho$ is set to 0.8.
Following \citet{conll2002}, we use the entity level F1-score as evaluation metric.
Moreover, experiments are repeated 5 times for different random seeds on each corpus.

Note that the selected data from the discriminator is generated by combination of output from mBERT-TLADVs with different random seeds, as we observe only a small number of samples with high $\ell_{\t{score}}$ in the selected data generated by each model. 
Specifically, for each target-language sentence $x^\t{T}$, there are 5 corresponding soft label sequences generated from 5 different mBERT-TLADV models. From those, only sequences that have the highest sum of each predicted label confidence are kept.

Our models are trained on a Tesla P100 GPU (16GB). mBERT-TLADV has 178M parameters and trains in $\approx$130min, while the student models $h^{\t T}_{\t{stu}}$ have 177M parameters and take $\approx$21min.

\subsection{Comparison with State-of-The-Art Results}
\label{sec:Comparison}

\begin{table*}[!t]
   \centering
   \begin{tabular}{@{}l|cccc@{}}
   \toprule
    \textbf{Model} & \textbf{de} & \textbf{es} & \textbf{nl} & \textbf{Avg}\\ \midrule
		\citet{tackstrom2012} & 40.40 & 59.30 & 58.40 & 52.70 \\
		\citet{tsai2016cross} & 48.12 & 60.55 & 61.56 & 56.74 \\
		\citet{ni2017weakly} & 58.50 & 65.10 & 65.40 &	63.00	 \\ 
		\citet{mayhew2017cheap} & 57.23 & 64.10 & 63.37 & 61.57 \\ 
		\citet{xie-etal-2018-neural} & 57.76 & 72.37 & 71.25 & 67.13 \\ 
		\citet{jain2019entity} & 61.5  & 73.5  & 69.9 & 68.30  \\ 
		\citet{bari2019zero} & 65.24 & 75.93 & 74.61 & 71.93 \\ 
		\citet{wu-dredze-2019-beto} & 69.56 & 74.96	& 77.57 & 73.57 \\ 
		\citet{adv-follow} & 71.9 & 74.3  & 77.6 & 74.60 \\
		\citet{wu2020enhanced} & 73.16 & 76.75 & 80.44	& 76.78 \\
		\citet{wu2020unitrans}* & 73.61 $\pm$ 0.39 & \underline{77.3} $\pm$ 0.78 & \underline{81.2} $\pm$ 0.83 & \underline{77.37} $\pm$ 0.67 \\ 
      \hline
       mBERT-ft & 72.59 $\pm$ 0.31 & 75.12 $\pm$ 0.83 & 80.34 $\pm$ 0.27 & 76.02 $\pm$ 0.47 \\ 
       mBERT-TLADV & \underline{73.89} $\pm$ 0.56 & 76.92 $\pm$ 0.62 & 80.62 $\pm$ 0.56 & 77.14 $\pm$ 0.58 \\ 
       AdvPicker & \textbf{75.01} $\pm$ 0.50 & \textbf{79.00} $\pm$ 0.21 & \textbf{82.90} $\pm$ 0.44 & \textbf{78.97} $\pm$ 0.38 \\ 
    \bottomrule
   \end{tabular}%
   \caption{Results of our approach and prior state-of-the-art methods for zero-shot cross-lingual NER. * denotes the version of the method without additional data.}
   \label{tab:main_results}
\end{table*}

Table \ref{tab:main_results} reports the zero-shot cross-lingual NER results of different methods on the 3 target languages. 
These include AdvPicker, prior SOTA methods, and two re-implemented baseline methods, i.e., mBERT-TLADV (Section \ref{sec:tladv}) and mBERT-ft (mBERT fine-tuned on labeled source-language data).
Note that some existing methods use the translation model as an additional data transfer source, whereas our method does not.
For a fair comparison, we compare against the version of UniTrans \citep{wu2020unitrans} w/o translation (as reported in their paper).
Our method outperforms the existing methods with F1-scores of 75.01, 79.90, and 82.90, when using only source-language labeled data and target-language unlabeled data.
Particularly, compared with Unitrans* (previous SOTA), AdvPicker achieves an improvement of F1-score ranging from 1.41 in German to 1.71 in Dutch. 
Furthermore, our result is comparable to the full UniTrans using also translation (0.04 F1 difference on average).

Besides, AdvPicker achieves an average F1-score improvement of 1.83 over mBERT-TLADV and 2.95 over mBERT-ft. These results well demonstrate the effectiveness of the proposed approach, which is mainly attributed to more effectively leveraging unlabeled target language data and selecting the language-independent data for cross-lingual transfer.

\subsection{Quality of Selected Data}
\label{sec:Quality}

\begin{table}[!t]
   \centering
   \vskip 0.1in
   \begin{tabular}{@{}l|cccc@{}}
   \toprule
      & \textbf{de} & \textbf{es} & \textbf{nl} & \textbf{Avg} \\ \midrule
	mBERT & 99.08 & 99.66 & 98.98 & 99.24  \\
	mBERT-ft & 98.38 & 98.66 & 97.27 & 98.10  \\
	mBERT-TLADV & 79.62 & 82.89 & 77.45 & 79.99  \\
    \bottomrule
   \end{tabular}%
   \caption{Accuracy of discriminators for different models in the three target languages.}
   \label{tab:discriminator_result}
\end{table}
   
\begin{table}[!t]
   \centering
   \vskip 0.1in
   \begin{tabular}{@{}l|ccc@{}}
   \toprule
    \textbf{Language} & \textbf{Selected} & \textbf{Other} & \textbf{$\Delta$} \\ \midrule
		de & 77.87 & 63.83 & 14.04 \\
		es & 76.45 & 69.23 & 7.22 \\
		nl & 84.33 & 66.97 & 17.36 \\
		Avg & 79.55 & 66.68 & 12.87 \\
    \bottomrule
   \end{tabular}%
   \caption{Teacher model F1 scores over target language training sets, w/o adversarial approach and distillation.}
   \label{tab:ablation_selected_f1}
\end{table}
   
\begin{table}[!t]
   \centering
   \vskip 0.1in
   \begin{tabular}{@{}l|cccc@{}}
   \toprule
    \textbf{Data type} & \textbf{de} & \textbf{es} & \textbf{nl} & \textbf{Total} \\ \midrule
		Selected data & 9733 & 6724 & 12668 & 29125 \\
		Other data & 2434 & 1681 & 3168 & 7283 \\
    \bottomrule
   \end{tabular}%
   \caption{
   Sentence numbers of the Selected/Other splits in the target language training sets.
   }
   \label{tab:ablation_selected_sentence}
\end{table}

\noindent \textbf{Language-Independence} We use the selected dataset $\mathbb{D}_{\t{subset}}$ to train student models via knowledge distillation. 
$\mathcal{P}_{\theta}({\v Y}^{\t{T-NER}})$ in $\mathbb{D}_{\t{subset}}$ is calculated over feature vectors generated by mBERT-TLADV. 
To validate the language-independence of these feature vectors, we apply three discriminators defined as in Eq.(\ref{equ:discrminator}) to classify the token feature vectors from three different encoders: mBERT, mBERT-ft, and mBERT-TLADV. 

Unlike in the adversarial learning setting, we fix the parameters of the three encoders and only train the discriminators. 
We use each language training set to train discriminators and evaluate on each target language corresponding test set. 
Table \ref{tab:discriminator_result} reports the discriminator accuracy for 3 different encoders.
We can see that the classification accuracy is reduced with adversarial training, which suggests that the similarity of feature vectors between the source language and the target language is improved; which further demonstrates the feature vectors become more language-independent when adversarial training is applied.

\begin{table*}[!t]
   \centering
   \vskip 0.1in
   \begin{tabular}{@{}l|cccc@{}}
   \toprule
    \textbf{Model} & \textbf{de} & \textbf{es} & \textbf{nl} & \textbf{Avg} \\ \midrule
		AdvPicker  & 75.02 & 79.00 & 82.90 & 78.97 \\
		\hline
		mBERT-ft & 72.59 (-2.43) & 75.12 (-3.88) & 80.34 (-2.56) & 76.02 (-2.95) \\
		mBERT-TLADV & 73.89 (-1.13) & 76.92 (-2.08) & 80.62 (-2.28) & 77.14 (-1.83) \\
		AdvPicker w/o KD & 73.98 (-1.04) & 77.91 (-1.09) & 80.55 (-2.35) & 77.48 (-1.49) \\
		AdvPicker w All-Data & 74.02 (-1.00) & 78.72 (-0.28) & 80.69 (-2.21) & 77.81 (-1.16) \\
    \bottomrule
   \end{tabular}%
   \caption{Ablation study for the proposed \textit{AdvPicker}, where numbers in parenthesis denote performance change.}
   \label{tab:ablation_result}
\end{table*}

\noindent \textbf{Pseudo Labels} To evaluate the language-independent quality of pseudo labels, we calculate the F1 scores of the pseudo labels and the number of sentences involved. 
We denote language-independent data and language-specific data as ``selected data" and ``other data" respectively.

Table \ref{tab:ablation_selected_f1} reports the pseudo labels F1 scores of both language-independent data and language-specific data for each target language using mBERT-TLADV. 
Generally, the average F1 score of language-independent data is 12.87 points higher than language-specific data, which suggests that language-independent data has higher quality pseudo-labels.
Furthermore, the selected data contains less language-specific information.

Table \ref{tab:ablation_selected_sentence} reports the number of language-independent and language-specific examples for each language.
From these results (Tables \ref{tab:ablation_selected_f1}, \ref{tab:ablation_selected_sentence}), we observe that the selected data is still high-quality, even if we set a very loose threshold (80\% of unlabeled data being selected).

\subsection{Model Performance over Selected/Other Data Splits}
\label{sec:selected-other}
\red{\begin{table*}[!t]
   \centering
   \vskip 0.1in
   \begin{tabular}{@{}l|ccccccc@{}}
   \toprule
    \textbf{Methods} & \textbf{de (Selected)} & \textbf{de (Other)} & \textbf{es (Selected)} & \textbf{es (Other)} & \textbf{nl (Selected)} & \textbf{nl (Other)} \\ \midrule
		mBERT-ft & 73.65 & 70.66 & 77.29 & 70.39 & 81.67 & 69.89 \\
		mBERT-TLADV & 74.05 & 72.49 & 78.04 & 73.86 & 81.83 & 77.89 \\
		UniTrans w/o translation & 74.48 & 71.71 & 77.29 & 73.18 & 83.15 & 70.39 \\
		AdvPicker & 75.11 & 73.76 & 79.19 & 75.68 & 84.19 & 79.15 \\
    \bottomrule
   \end{tabular}%
   \caption{F1 scores over the Select/Other test set splits in different target languages. \textit{AdvPicker} has better performance also on Other data for all languages.}
   \label{tab:test_data_split_result}
\end{table*}}

In order to better analyse the behaviour of \textit{AdvPicker} across data variations, we use the trained language discriminator to split the target language test sets into Selected and Other partitions (similarly to how the training set is processed). Table \ref{tab:test_data_split_result} shows the different models's F1 scores for the partitioned data.

From Table \ref{tab:test_data_split_result}, we can draw these conclusions:

1) As expected, models perform better over Selected data than over Other data;

2) \textit{AdvPicker} is only trained on Selected data, but nonetheless outperforms all baseline models in both data partitions;

3) \textit{AdvPicker}'s approach effectively selects examples with better features and is not over-biased towards Selected data.

\subsection{Ablation Study}
\label{sec:Ablation}

To validate the contributions of different process in the proposed \textit{AdvPicker}, we introduce the following variants of our approach and baselines to perform an ablation study:
1) \textit{AdvPicker w/o KD}, which directly combines the prediction of test data from \textit{mBERT-TLADVs} with different seeds without knowledge distillation on pseudo-labeled training data.
2) \textit{AdvPicker w All-Data}, which trains a student model on all target-language pseudo-labeled data generated by \textit{mBERT-TLADV}.
3) \textit{mBERT-ft}, mBERT fine-tuned on source-language labeled data.
4) \textit{mBERT-TLADV} (Section \ref{sec:tladv}), meaning mBERT trained on source-language labeled data with token-level adversarial learning.

Table \ref{tab:ablation_result} reports the performance of each method and their performance drops compared to AdvPicker. Moreover, we can draw more in-depth observations as follows:

1) Comparing \textit{AdvPicker} with \textit{AdvPicker w/o KD} and \textit{AdvPicker w All-Data}, we can see that selecting the language-independent data is reasonable.
That also validates the effectiveness of the model trained on language-independent data via knowledge distillation.

2) \textit{mBERT-ft} outperforms \textit{mBERT-TLADV}.
Such results well demonstrate that token-level adversarial learning is helpful to train a language-independent feature encoder and brings performance improvement.  

3) By comparing the F1 scores of \textit{AdvPicker w All-Data} and \textit{AdvPicker} on the target languages, we observe that training on selected data brings higher performance improvements on larger datasets, e.g., German [de] and Dutch [nl], and lower improvements on the smaller Spanish [es] dataset. 
Although selected data has high-quality pseudo labels, smaller sizes of selected datasets may limit performance improvements. 

\subsection{Stability Analysis}
\label{sec:Stability}
Because BERT fine-tuning is known to be unstable in few-shot tasks, as discussed in \citet{unstable}.
mBERT-based methods' performances on the CoNLL NER dataset are likely also unstable.
To evaluate the stability of AdvPicker, we compare the standard deviation of F1 scores for mBERT-ft, Unitrans, and AdvPicker.

Table \ref{tab:main_results} includes the standard deviation of F1 scores over five runs for each model.
\textit{AdvPicker} has a lower average standard deviation in the three target languages than the other mBERT-based methods. Such results demonstrate that selected data can bring a degree of stability to the model, or limit instability, as the student model in \textit{AdvPicker} is trained on selected data with the soft labels from other trained models. 

\section{Conclusion}
\label{sec:Conclusion}
In this paper, we propose a novel approach to combine the feature-based method and pseudo labeling via language adversarial learning for cross-lingual NER. \textit{AdvPicker} is the first successful attempt in selecting language-independent data by adversarial discriminator to cross-lingual NER. Our experimental results show that the proposed system benefits strongly from this new data selection process and outperforms existing state-of-the-art methods, even without requiring additional extra resources.

\bibliographystyle{unsrtnat}
\bibliography{XL-Adv}

\end{document}